%% file: main.tex
\documentclass[10pt,twocolumn,letterpaper]{article}

\usepackage[pagenumbers]{cvpr} % To force page numbers, e.g. for an arXiv version

\input{preamble}
\definecolor{cvprblue}{rgb}{0.21,0.49,0.74}
\usepackage[pagebackref,breaklinks,colorlinks,allcolors=cvprblue]{hyperref}
\usepackage{multirow}
\usepackage{placeins}
\def\paperID{12596} % *** Enter the Paper ID here
\def\confName{CVPR}
\def\confYear{2026}

\title{Portable Active Learning for Object Detection}

\author{
 Rashi Sharma$^{1}$ \quad Justin Timothy C. Bersamin$^{2}$\thanks{Work done while an intern at Panasonic R\&D Center Singapore.
See author contributions  for further details.} \quad Karthikk Subramanian$^{1}$ \\
 $^{1}$Panasonic R\&D Center Singapore \quad $^{2}$Nanyang Technological University \\
 {\tt\small rashi.sharma@sg.panasonic.com, bers0001@e.ntu.edu.sg, karthikk.subramanian@sg.panasonic.com}
}

\begin{document}
\maketitle
\input{sec/0_abstract}
\input{sec/intro}

\input{sec/relwrk}
\input{sec/method}

\input{sec/Experiment}

\input{sec/conclusion}

\input{sec/ac}

{
    \small
    \bibliographystyle{ieeenat_fullname}
    \bibliography{main}
}

% WARNING: do not forget to delete the supplementary pages from your submission 
% \input{sec/X_suppl}

\end{document}

%% file: sec/0_abstract.tex
\begin{abstract}
Annotating bounding boxes is costly and limits the scalability of object detection. This challenge is compounded by the need to preserve high accuracy while minimizing manual effort in real-world applications. Prior active learning  methods often depend on model features or modify detector internals and training schedules, increasing integration overhead. Moreover, they rarely jointly exploit the benefits of image-level signals, class-imbalance cues, and instance-level uncertainty for comprehensive selection. We present Portable Active Learning (PAL), a detector-agnostic, easily portable framework that operates solely on inference outputs. PAL combines class-wise instance uncertainty with image-level diversity to guide data selection. At each round, PAL trains lightweight class-specific logistic classifiers to distinguish true from false positives, producing entropy-based uncertainty scores for proposals. Candidate images are then refined using global image entropy, class diversity, and image similarity, yielding batches that are both informative and diverse. PAL requires no changes to model internals or training pipelines, ensuring broad compatibility across detectors. Extensive experiments on COCO, PASCAL VOC, and BDD100K demonstrate that PAL consistently improves label efficiency and detection accuracy compared to existing active learning baselines, making it a practical solution for scalable and cost-effective deployment of object detection in real-world settings.

\end{abstract}

%% file: sec/intro.tex
\section{Introduction}
The rapid progress in deep learning has led to remarkable advances in object detection, enabling robust performance across a wide range of visual tasks. However, the success of these models is heavily dependent on the availability of large-scale, high-quality annotated datasets, particularly those with precise bounding box labels. The process of annotating such data is both time-consuming and expensive, often becoming a bottleneck in deploying object detection systems for new domains or rare categories.

Active learning (AL) has emerged as a promising paradigm to address this challenge by intelligently selecting the most informative samples for annotation, thereby reducing the overall labelling effort required to achieve strong model performance. While AL has shown significant benefits in image classification \cite{ClassBalancedAL, ThePO, aal}, its application to object detection introduces unique challenges. These include handling uncertainty caused by both localization and classification, managing intra-class imbalance within an image impacting image uncertainty, and ensuring that the selected samples contribute to both uncertainty reduction and dataset diversity.

Existing approaches to active learning for object detection often  require modifications to the training pipeline or model internals \cite{mial, mdn, ll, ppal}, which can limit their generalizability and ease of adoption.
Moreover, existing methods tackling AL for detection with instance uncertainty \cite{mial, ll, sraal}, rarely jointly exploit image-level signals and class-imbalance cues with instance-level uncertainty to construct an overall informative and diverse annotation batch.

\begin{figure*}[t]
  \centering
  %\fbox{\rule{0pt}{2in} \rule{0.9\linewidth}{0pt}}
   \includegraphics[width=1.0\linewidth]{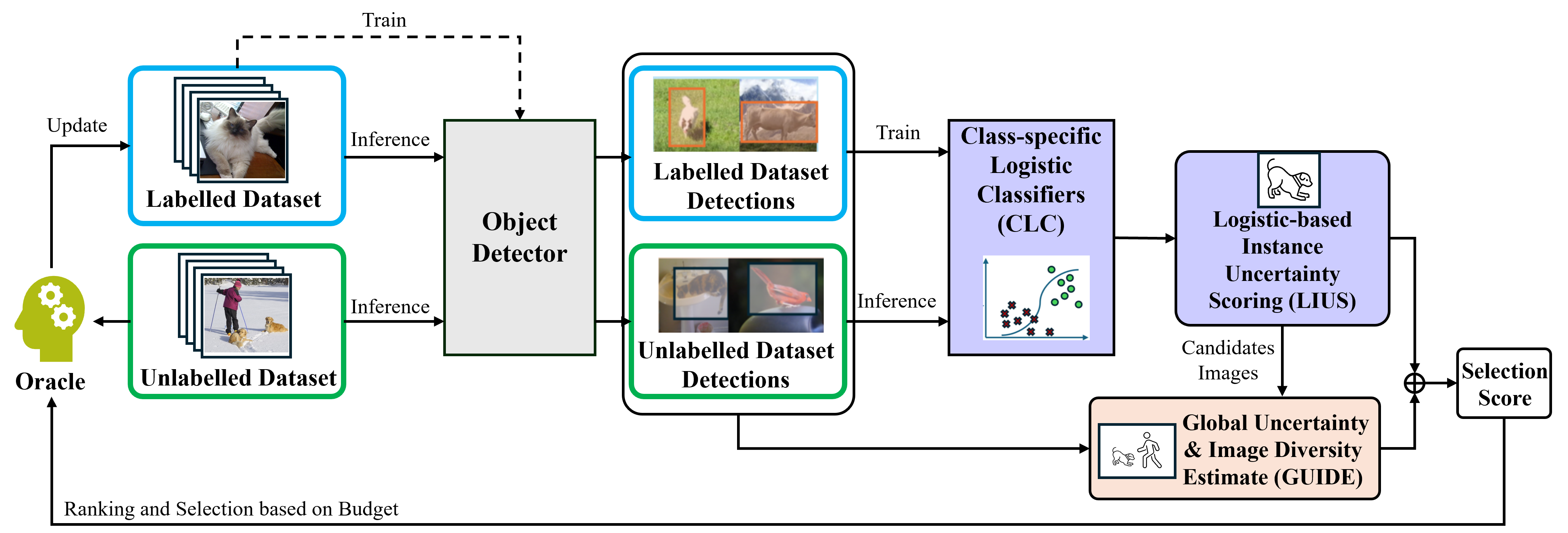}

   \caption{Architectural overview of our Portable Active Learning (PAL) framework. The object detector, trained on the current iteration's annotated dataset, generates predictions for both labelled and unlabelled image sets. Detections from the labelled set are utilized to train CLC for LIUS. The GUIDE score incorporates: class-weighted image entropy, rare-class driven diversity index, and rank-conditioned similarity penalty. LIUS and GUIDE scores are combined to determine the optimal subset of images for annotation by the oracle within the specified budget constraints for next training round.}
   \label{fig:overall}
   \vspace{-10pt}
\end{figure*}

We propose Portable Active Learning (PAL), a detector and training-pipeline agnostic, easily portable active learning framework that operates solely on inference outputs making it easy to generalize and deploy across detectors and domains. PAL couples class-wise instance uncertainty computed using logistic-based instance uncertainty scoring (LIUS) with complementary image-level signals—class-weighted image entropy, rare-class driven diversity index, and rank-conditioned similarity penalty— aggregated into a global uncertainty and image diversity estimate (GUIDE). This ensures uncertainty reduction and dataset diversity while targeting class imbalance. In LIUS, PAL prioritizes images containing objects for which the detector is most uncertain by training lightweight class-specific logistic classifiers (CLC) to distinguish true positives from false positives using only two detection-derived features: pre-NMS box count and detection confidence. The image-level signals of GUIDE further ensure that selected images are globally informative rather than narrowly focused on a single class instance. By integrating instance uncertainty, image-level uncertainty and diversity, and a low-frequency class aware class budgeting mechanism, PAL selects annotation batches that are informative, diverse, and balanced.

The contributions of this paper include:

\textbf{(1)} We introduce PAL, a portable AL framework for object detection consisting of a two-part scoring algorithm that fuses instance-level and image-level signals to provide a comprehensive measure of image informativeness and diversity.

\textbf{(2)} We show PAL's model-internals and training-pipeline independent capabilities that enable broad compatibility and simple integration with diverse detectors. 

\textbf{(3)} We demonstrate consistent gains in label efficiency and detection accuracy over active learning baselines across multiple object detectors on COCO, PASCAL VOC, and BDD100K datasets.

%% file: sec/relwrk.tex
\section{Related Work}

%Active learning for object detection has attracted significant attention due to its potential to reduce annotation costs while maintaining high detection accuracy. 
Existing active learning approaches can broadly be categorized into uncertainty-based methods, diversity-driven strategies and hybrid approaches.

%Early work in AL for detection primarily adapted uncertainty sampling to the localization setting. \cite{laal} introduced localization-aware metrics (tightness, stability). \cite{burst} combined entropy with feature clustering, emphasizing the need for diversity beyond uncertainty alone.

\textbf{Uncertainty-Based Selection:}
Early works in active learning for object detection primarily focused on uncertainty estimation at the instance level. Common techniques include entropy-based scoring, margin sampling, and Bayesian approximations \cite{uncertainty1, uncertainty2}, which aim to identify predictions with high ambiguity. While effective in classification tasks, these methods often struggle in detection scenarios due to the presence of multiple proposals per image and the inherent noise in bounding box predictions. Methods such as MIAL~\cite{mial} extend uncertainty-based selection by learning instance-level uncertainty via adversarial classifiers and aggregating image-level uncertainty through multiple instance learning, effectively filtering out background and noisy instances. Another notable direction is the use of loss prediction modules, as in Learning Loss for Active Learning~\cite{ll}, which estimates the expected loss for each sample to guide selection. This approach has shown promise in classification and detection tasks, but requires additional network components and training overhead.

\textbf{Diversity and Representativeness:}
To complement uncertainty, several methods incorporate diversity measures to ensure that selected samples cover a broad range of visual and semantic variations. Strategies include clustering in feature space, core-set selection, and adversarial sampling \cite{coreset, adversarial}. However, these techniques often require access to intermediate model representations or gradients, which complicates integration with off-the-shelf detectors. Moreover, balancing diversity with uncertainty remains a challenge, as naive combinations can lead to redundant or uninformative selections. To address class imbalance and ensure rare categories are adequately represented, contextual diversity~\cite{cdal} and class-balanced active learning~\cite{ClassBalancedAL} have been proposed. These methods explicitly promote the selection of samples from under-represented classes, which is particularly important for real-world datasets with long-tailed distributions. Despite these advances, achieving an effective balance between diversity, uncertainty, and class representation remains an open challenge, motivating the development of hybrid and more practical approaches.

\textbf{Hybrid Approaches:}
Recent research has explored combining uncertainty with diversity-aware selection to improve robustness. Methods such as PPAL~\cite{ppal} and DivProto~\cite{divproto} introduce multi-stage scoring pipelines that first select candidate samples based on uncertainty and then refine the selection using diversity or representativeness measures. PPAL, for example, uses difficulty-calibrated uncertainty sampling followed by category-conditioned matching similarity and k-means++ clustering to ensure both informativeness and diversity in the selected batch. Semi-supervised and large-scale active learning approaches~\cite{softteacher, activeteacher} further leverage unlabelled data and pseudo-labelling to reduce annotation costs. Despite these advances, practical deployment of active learning in object detection remains challenging due to the need for model-agnostic, scalable, and easy-to-integrate solutions. In contrast, our proposed framework, PAL, is fully inference-driven and detector-agnostic. By leveraging logistic-based instance uncertainty scoring (LIUS) and a global uncertainty and image diversity estimate (GUIDE), PAL addresses both uncertainty and representativeness in a unified manner.

%logistic regression models trained on labeled data to estimate true/false positive boundaries, PAL computes entropy-based uncertainty without modifying the detector. Furthermore, PAL integrates global image entropy, class diversity, and image similarity to refine selection, addressing both uncertainty and representativeness in a unified manner.

%% file: sec/method.tex
\section{Method}
\label{sec:method}
This section is divided into the following: problem of active learning for object detection \cref{sec:ps}, key innovations of our method detailed in \cref{sec:pmo}. 

%-------------------------------------------------------------------------
\begin{figure*}[t]
  \centering
  %\fbox{\rule{0pt}{2in} \rule{0.9\linewidth}{0pt}}
   \includegraphics[width=1.0\linewidth]{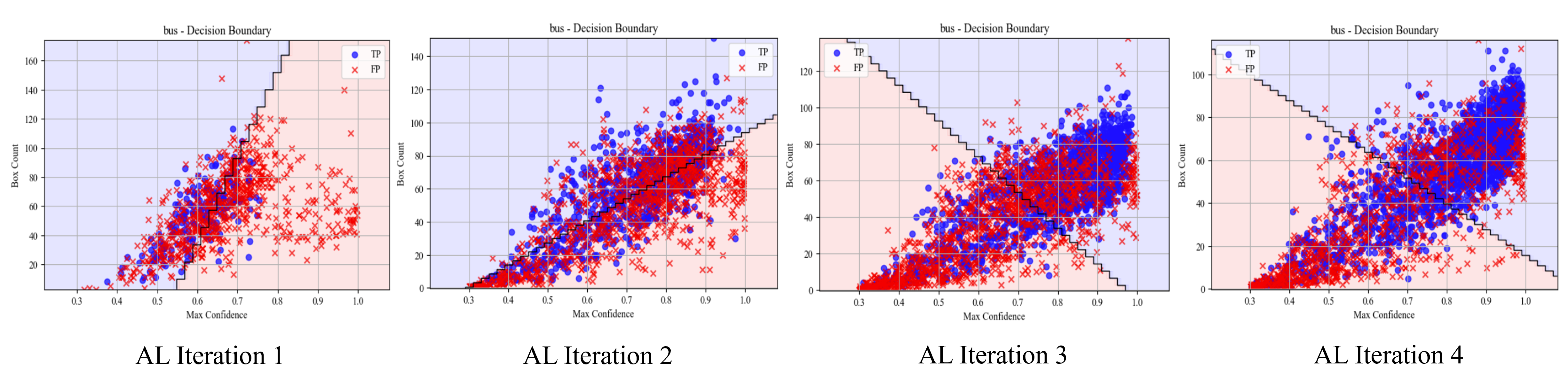}
\caption{PAL’s class-specific logistic classifiers (CLC) across AL data selection rounds (1–4) for the bus category, a low-frequency class in the BDD100K dataset. Each CLC graph has max detection confidence on X-axis and pre-NMS box counts per detection on Y-axis. In iteration 1, trained on an initial random data, the detector detects 866 bus samples (675 false positives (FP), 191
true positives (TP)), used to train CLC on bus class. At this stage, TP and FP regions are largely inseparable in the learned feature space
with many FPs shifted towards high confidence region. In subsequent iterations as PAL adds more TP and FP instances of bus class, the
difference between TP/FP regions becomes more apparent: FPs move toward low confidence areas while TPs shift toward high confidence
and high box count region, indicating improved detector performance on bus class.}
   \label{fig:lr}
   \vspace{-10pt}
\end{figure*}
\subsection{Problem Statement}
\label{sec:ps}
Active learning aims to minimize the effort of human annotation while maximizing model performance by intelligently selecting the most informative samples for labelling. In the context of object detection, we formulate active learning as an iterative process consisting of three key phases: model training, sample selection, and oracle annotation.

Formally, let $ U = \{u_i\}_{i \in [N]} $ denote a large pool of unlabelled images, where N is the size of the unlabelled dataset, and $ L = \{l_i\}_{i \in [M]} $ represent an initially labelled set of M randomly chosen images, such that $U \cap L = \emptyset$. The active learning problem can be defined as follows:
In each active learning round r ($r > 0$), we aim to select a subset S such that $S_r \subseteq U_r$ of size b (where b is the annotation budget) that maximizes the expected performance gain of the detector. After oracle annotation of $S_r$, the labelled and unlabelled sets are updated as: 
\begin{equation}
\begin{split}
L_{r+1} = L_r \cup S_r, \\
U_{r+1} = U_r - S_r.
\label{eq:al}
\end{split}
\end{equation}
The detector is then retrained on the expanded labelled set $L_{r+1}$. This process continues iteratively, with the objective of achieving optimal detection performance on a predefined validation set while minimizing the total annotation cost. The effectiveness of an active learning method is measured by its ability to achieve higher detection performance with fewer labelled samples compared to other baseline strategies.

%-------------------------------------------------------------------------
\subsection{Proposed Method Overview}
\label{sec:pmo}
The high-level overview architecture of PAL is provided in \cref{fig:overall}. PAL's data selection for active learning consists of a two-part scoring function that combines instance-level and image-level scores. At instance level we introduce a novel method, Logistic-based Instance Uncertainty Scoring (LIUS) to quantify the model’s uncertainty regarding the existence of each predicted instance. Details about LIUS are provided in \cref{sec:is}.

Our image-level scoring function, called the Global Uncertainty and Image Diversity Estimate (GUIDE), is an aggregate of three cues: Class-Weighted Image Entropy (CWIE), Rare-Class driven Diversity Index (RCDI), and Rank-Conditioned Similarity Penalty (RCSP). Additional details of GUIDE appear in \cref{sec:gis}.

The two components are summed into a final selection score for a given instance j in image I by:
\begin{equation}
  Score(I,j) =  \alpha\cdot S_{LIUS}(I_j) +  d\cdot S_{GUIDE}(I),
  \label{eq:ts}
\end{equation}
where, $\alpha+d=1$. $S_{LIUS}(I_j)$ refers to the instance-level score, $S_{GUIDE}(I)$ is the image-level score, and $\alpha$ and $d$ are weights of contribution of the two scores respectively, towards the final score. More details in \cref{sec:tis}. Given budget b, using selection score we select the top-scoring images for annotation by the oracle, for the next training round. 

\subsubsection{Logistic-based Instance Uncertainty Scoring}
\label{sec:is}
Logistic-based Instance Uncertainty Scoring (LIUS) is a novel instance-level uncertainty scoring mechanism that leverages pre-NMS box counts and box confidence scores to identify ambiguous detections. Our approach operates in two phases: (1) learning a decision boundary between true and false positives using labelled dataset features, and (2) calculating uncertainty of unlabelled instances based on their proximity to this boundary.

First, we perform inference on both labelled and unlabelled datasets, extracting three key features for each detection: the number of pre-NMS boxes, detection confidence, and predicted class. For labelled data, we additionally record the ground truth status (true/false positive) of each detection. The pre-NMS box count serves as a crucial feature, as dense clusters of high-confidence pre-NMS boxes often correlate with object presence. Pre-NMS box count is calculated by first extracting pre-NMS boxes of the entire image. After NMS yields final detections, the pre-NMS boxes are assigned to each detection using an IoU threshold. The number of boxes assigned defines the pre-NMS count for that detection.

For each class c, we train a binary logistic classifier (CLC from \cref{fig:overall}) on the labelled data, represented by:
\begin{equation}
 P(Y=1 | \mathbf{x}) = \frac{1}{1 + e^{-(\beta_0 + \beta_1 x_1 + \beta_2 x_2)}}.
  \label{eq:lrc}
\end{equation}
Here $x_1$ and $x_2$ represent the pre-NMS box count and detection confidence, respectively, and $P(Y=1 | \mathbf{x})$ represents the probability of the instance being a true positive given the features $\mathbf{x}$. The classifier learns to estimate the decision boundary between true and false positives in the feature space.

We then use these class-specific logistic classifiers to infer the probability score of the unlabelled pool instances. Given the probability score $P(Y_j=1 | \mathbf{x_j})$ of instance j from image I, we compute its uncertainty using Shannon Entropy, which acts as the LIUS score:
%H(d) = -p(\text{TP}|d)\log(p(\text{TP}|d)) - (1-p(\text{TP}|d))\log(1-p(\text{TP}|d))
\begin{equation}
LIUS(I_j) = - \sum_{Y_j \in (0,1)} P(Y_j) \log P(Y_j).
  \label{eq:se}
\end{equation}
The most informative instances are selected according to their LIUS score. Across datasets, it is observed that frequently occurring classes tend to dominate the instance selection based budget. To improve the frequency of under-represented classes, we introduce class based budgeting in which we allocate the annotation budget b across classes c according to a weighted ratio $r_c$ calculated by:
\begin{equation}
r_c = 1-0.5\cdot(n_{c,l}/N_l + n_{c,u}/N_u),
  \label{eq:rc}
\end{equation}
where $n_{c,l}$ and $n_{c,u}$ are the instance counts of class c in labelled and unlabelled sets respectively, and $N_l$, $N_u$ is the total number of labelled and unlabelled dataset detections. Given $r_c$ each class c, class budget $b_c$ is assigned by:
\begin{equation}
b_c = min(n_{c,u}, b\cdot(\frac{r_c}{\sum_{ci \in C} r_{ci}})),
  \label{eq:bc}
\end{equation}
such that,
$$\sum_{c \in C} b_c = b, $$
where, C represents all classes, b is total budget and $r_{ci}$ is $r_c$ from \cref{eq:rc} summed over all classes. This class budgeting method favours minority classes with hopes of improving their presence in labelled dataset. For each class c, we select the top $2\cdot b_c$ candidate images containing the highest LIUS  instances for that class. These candidate images proceed to GUIDE for final selection.
LIUS effectively identifies regions in data distribution where the model exhibits maximum uncertainty in distinguishing true from false detections in the classification feature space, prioritizing them for annotation. As more instances from this distribution are selected, the detector's ability to identify class features pertinent to detect the class improves, as seen in \cref{fig:lr}.

%\begin{figure}[b]
%  \centering
%  %\fbox{\rule{0pt}{2in} \rule{0.9\linewidth}{0pt}}
%   \includegraphics[width=0.95\linewidth]{sec/figs/gos.png}
%
%    \caption{GUIDE score components.}
%   \label{fig:glbs}
%\end{figure}

\subsubsection{Global Uncertainty \& Image Diversity Estimate}
\label{sec:gis}
\begin{figure*}[t]
  \centering
  %\fbox{\rule{0pt}{2in} \rule{0.9\linewidth}{0pt}}
   \includegraphics[width=0.86\linewidth]{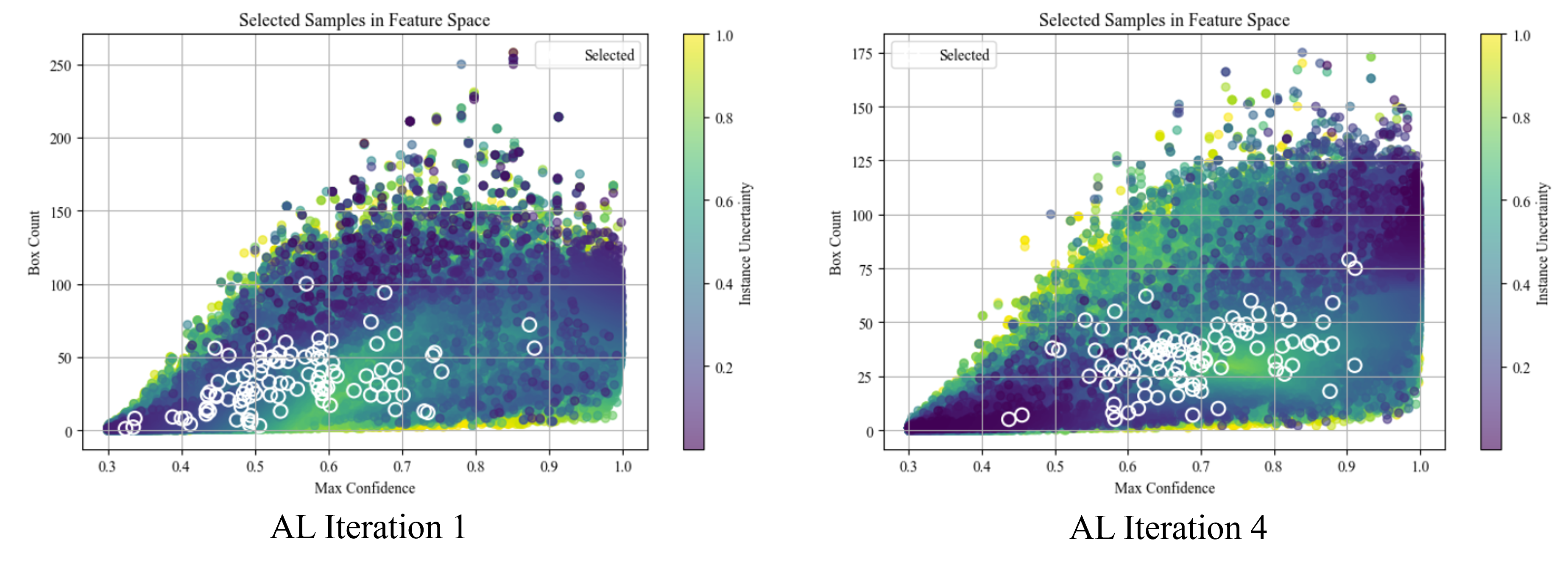}

   \caption{Feature space of PAL’s logistic classifier visualizing detections across classes on the COCO dataset. Marker color encodes LIUS score, and white borders highlight the top 100 detections selected for annotation. In iteration 1, selected instances cluster in regions of low detector confidence and few pre-NMS overlapping bounding boxes. By iteration 4, the high-entropy region and selected cluster has shifted toward the center of the feature space, reflecting the detector’s improved confidence of true positives. }
   \label{fig:fs}
   \vspace{-10pt}
\end{figure*}
%Prior works have encouraged diversity via feature-space clustering, core-set selection, gradient-based sampling, and adversarial or variational schemes []. Many successful approaches rely on features produced by the model. 
LIUS only focuses on choosing images with informative class objects while ignoring the overall informativeness of the images containing those chosen objects. To remedy this, we inject Global Uncertainty and Image Diversity Estimate (GUIDE) into the equation to improve the image selection by involving image-level signals. However, to make PAL image selection independent of model internals and training code, we leverage model-agnostic, image-level uncertainty and diversity measures that capture information useful for selecting images. In this second part of PAL, the selection of $b_c$ images per class is driven by three image-level signals: (1)  Class-Weighted Image Entropy (CWIE), (2) Rare-Class driven Diversity Index (RCDI), and (3) Rank-Conditioned Similarity Penalty (RCSP).
\\
\textbf{Class-Weighted Image Entropy (CWIE):} CWIE measures image level uncertainty while balancing the contribution of frequent versus rare classes. For image I with O detected objects, let $p_{i}$ denote the predicted confidence and c denote the class assigned to object i while $r_{c_i}$ from \cref{eq:rc} be the weight for class c. CWIE for I, is computed as:
\begin{equation}
CWIE(I) = - \sum_{i \in O} r_{c_i} \cdot \sum_{j \in C} p_{ij} \log p_{ij},
\label{eq:ei}
\end{equation}
where C is the number of classification ways. Class weights prevent heavily occurring classes from dominating the entropy signal. 

CWIE is normalized between [0, 1] on the $2\cdot b_c$ candidates, using min-max scaling: 
\begin{equation}
X_{std} = \frac{X - X_{min}}{X_{max} - X_{min}},
  \label{eq:std}
\end{equation}
where, $X_{min}$ is set to 0 and $X_{max}$ equals to the maximum image entropy score observed across the $2\cdot b_c$ candidates from instance scoring stage.
\newline
\textbf{Rare-Class driven Diversity Index (RCDI):} While CWIE reflects the number of informative objects in an image, it can be inflated by many instances of a dominant class. To favour images that contain informative objects spanning multiple especially rare classes, we compute RCDI per image I by:
\begin{equation}
RCDI(I) = \sum_{k \in K} r_k,
\label{eq:cdi}
\end{equation}
where, K is the unique classes in an image and $r_k$ is computed as per \cref{eq:rc}.  The same scaling as \cref{eq:std} is applied across all the candidates of each class, producing an RCDI score between [0, 1].
\\
\textbf {Rank-Conditioned Similarity Penalty (RCSP):} Prior works have encouraged diversity via feature-space clustering, core-set selection, gradient-based sampling, and adversarial or variational schemes \cite{coreset, adversarial, cdal, pcluster}. Other successful approaches rely on features produced by the model \cite{ppal} to calculate diversity. To make active learning data selection independent of model internals we rely only on image and inference level information for diversity. To achieve this PAL uses vision transformer-based image encoders \cite{vit} which convert images into low-dimensional embeddings, making it straightforward to measure pairwise image similarity via cosine similarity. In our case, we use a pre-trained, vision transformer encoder model \cite{gvit}, to compute embeddings for all images in the dataset. For each class, we rank the candidates by their LIUS score in descending order, i.e., higher the score, smaller the rank. The top-ranked image of each class is assigned a diversity score of 1. For each subsequent image ranked i, we compare its embedding $e_i$ one-to-one with the embeddings $e_m$, of all higher-ranked images with rank $m \in [1, i-1]$ using cosine similarity. RCSP of image I is computed by:
\begin{equation}
RCSP(I) = 1 - \max_{m \in [1, i-1]} (\cos(e_i, e_m)).
\label{eq:dsi}
\end{equation}
This ensures that if two images are similar, only the lower-ranked one is penalized. This prevents both images from being discarded based just on their similarity score with each other. Thus, the RCSP score enforces image-level diversity among class candidates, ensuring that selected high-entropy instances exhibit variation either in object appearance or in the surrounding context.
\begin{figure*}[t]
  \centering
  %\fbox{\rule{0pt}{2in} \rule{0.9\linewidth}{0pt}}
   \includegraphics[width=1.0\linewidth]{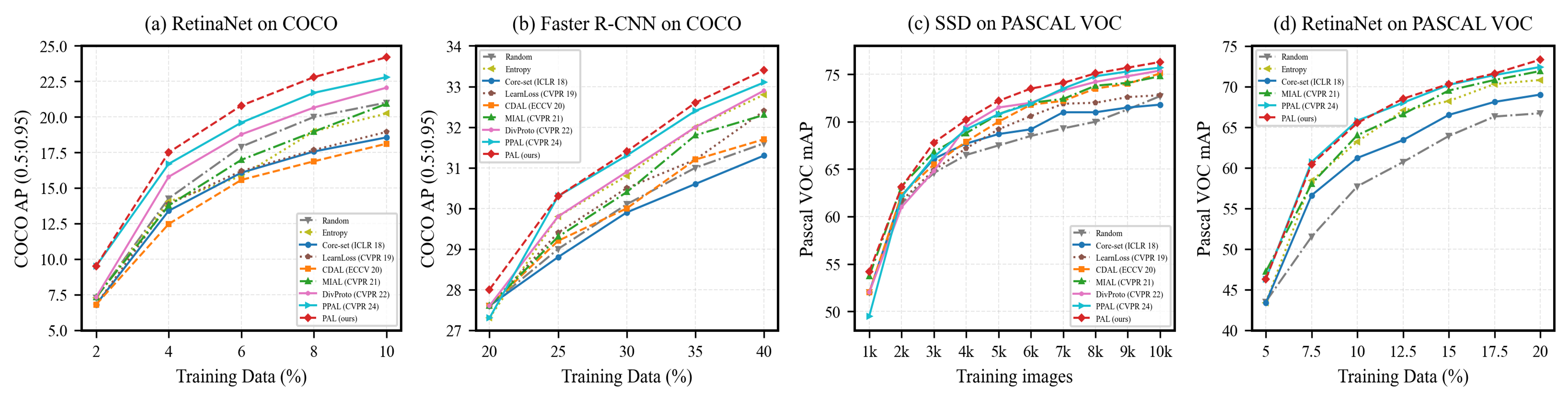}

   \caption{Comparison between PAL and state-of-the-art active learning algorithms. Subfigures (a) and (b) report AP (\%) on COCO using RetinaNet and Faster R-CNN, respectively, while (c) and (d) presents mAP (\%) on PASCAL VOC using SSD and RetinaNet, respectively.}
   \label{fig:res}
   \vspace{-10pt}
\end{figure*}
%\begin{figure}[t]
%  \centering
%  %\fbox{\rule{0pt}{2in} \rule{0.9\linewidth}{0pt}}
%   \includegraphics[width=0.75\linewidth]{sec/figs/rnvoc.png}
%
%   \caption{MAP(\%) results on PASCAL VOC using RetinaNet.}
%   \label{fig:vocres}
%\end{figure}
\subsubsection{Selection Score}
\label{sec:tis} 
After computing GUIDE scores via \cref{eq:ei}, \cref{eq:cdi} and \cref{eq:dsi} , the final per-image selection score for image I is computed as a weighted combination of LIUS and GUIDE scores:
\begin{equation}
\begin{split}
\text{Score(I)} = \alpha \cdot LIUS(I_j)+ \gamma \cdot RCSP(I) \\
+ \beta \cdot (CWIE(I) + RCDI(I)),
\label{eq:tsi}
\end{split}
\end{equation}

such that, $2\cdot \beta + \gamma = d$, where $\alpha$ and $d$ are parameters from \cref{eq:ts}. 
$LIUS(I_j)$ is from \cref{eq:se} where j is high LIUS object of a class due to which the image I was considered. GUIDE scores, RCSP(I) is from \cref{eq:dsi}, CWIE(I) is from \cref{eq:ei} and RCDI(I) is from \cref{eq:cdi}. $\gamma$ and $\beta$ are weights assigned to the individual components of GUIDE.

We keep weights same for CWIE and RCDI to balance the contribution of the number of informative boxes in an image to their class diversity. For each class, the top $b_c$ images ranked by their selection score are sent to the oracle for annotation.

%% file: sec/Experiment.tex
\section{Experiment}
\label{sec:exp}
\subsection{Experiment Settings}
\label{sec:es}
\textbf{Dataset Settings:}
We evaluated PAL on three standard object detection benchmarks: COCO \cite{coco}, PASCAL VOC \cite{voc}, and BDD100K \cite{bdd}.
\begin{itemize}
\item \textbf{COCO (80 classes)}: We trained on train2017 and evaluated on val2017 (5000 images). For RetinaNet \cite{retinanet}, we adopted the data protocol of MIAL \cite{mial}; for Faster R-CNN \cite{frcnn}, we followed the DivProto settings \cite{divproto}.
\item \textbf{PASCAL VOC (20 classes)}: We trained on the union of train2007 and train2012  and evaluated on test2007, consistent with \cite{ppal, mial}. The per-iteration data split for RetinaNet and SSD followed \cite{ppal}.
\item \textbf{BDD100K (BDD) (10 classes)}: We used the default split, with 70{,}000 training images and 10{,}000 validation images. For active learning, PAL initialized the labelled set with 2.5\% randomly selected training images and, at each cycle, annotated an additional 2.5\% from the remaining unlabelled pool for four rounds, reaching 12.5\% of the training set. The data settings were kept same across both RetinaNet and YOLOX-Tiny models \cite{yolox}.
\end{itemize}

 It is important to note that dataset configurations may vary depending on the detector used. To mitigate the effects of randomness, all experiments are repeated three times with different initial training sets, and the results are averaged.
\begin{table*}[t]
    \centering
      \resizebox{0.77\textwidth}{!}{
    \large
    \begin{tabular}{|c|c|c|c|c|c|c|}
        \hline
         \textbf{Model} & \textbf{Method} & \textbf{Round 1} & \textbf{Round 2} & \textbf{Round 3} & \textbf{Round 4} & \textbf{Round 5} \\
        %\multicolumn{2}{|c|}{\textbf{Group A}} & \multicolumn{4}{c|}{\textbf{Group B}} \\
        %\hline
        %\multirow{2}{*}{COCO} & \multirow{2}{*}{RetinaNet} & PPAL[] & $9.5\pm0.5$ & $16.7\pm0.1$ & $19.6\pm0.3$ & $21.7\pm0.1$ & $22.8\pm0.2$ \\
        %\cline{3-8}
        %& & Ours & $9.5\pm0.5$ & $\mathbf{17.5\pm0.4}$ & $\mathbf{20.8\pm0.3}$ & $\mathbf{22.8\pm0.3}$ & $\mathbf{24.2\pm0.2}$ \\
        \hline
        \multirow{4}{*}{RetinaNet} & Random & $26.8\pm0.8$ & $34.7\pm1$ & $37.8\pm0.6$ & $40.2\pm0.1$ & $42.2\pm0.2$ \\
        \cline{2-7}
        & Entropy & $26.8\pm0.8$ & $36.3\pm1.2$ & $41.5\pm0.4$ & $43.5\pm0.4$ & $44.8\pm0.1$ \\
        \cline{2-7}
        & PPAL & $26.8\pm0.8$ & $38.9\pm0.4$ & $42.5\pm0.3$ & $44.4\pm0.1$ & $45.5\pm0.3$ \\
        \cline{2-7}
        & Ours & $26.8\pm0.8$ & $\mathbf{40.1\pm0.5}$ & $\mathbf{43.7\pm0.2}$ & $\mathbf{45.7\pm0.2}$ & $\mathbf{46.7\pm0.2}$ \\
        
       \hline
        \multirow{3}{*}{YOLOX-Tiny} & Random & $9.9\pm0.3$ & $10.6\pm0.1$ & $11\pm0.3$ & $11.3\pm0.1$ & $11.5\pm0.2$ \\
        \cline{2-7}
        & Entropy & $9.9\pm0.3$ & $11.4\pm0.3$ & $11.6\pm0$ & $11.8\pm0.3$ & $12.2\pm0.1$ \\
        \cline{2-7}
        & Ours & $9.9\pm0.3$ & $\mathbf{12\pm0}$ & $\mathbf{12.6\pm0.2}$ & $\mathbf{13.1\pm0.1}$ & $\mathbf{13.3\pm0.2}$ \\
               
        %\multirow{2}{*}{Data Set 1} & \multirow{2}{*}{A1} & 10.5 & 2.1 & 75 & 1 \\
        %\cline{3-6}
        % & 9.8 & 3.0 & 82 & 0 \\
        %\hline
        %Data Set 2 & A2 & 15.0 & 1.5 & 60 & 1 \\
        \hline
    \end{tabular}
    }
    \caption{MAP(\%) results on BDD dataset using RetinaNet and YOLOX-Tiny models. The first round is random data round where 2.5\% of data was randomly selected from BDD. Subsequent rounds 2-5 are results from AL data selection for respective methods. In each AL round 2.5\% data is selected and annotated for training.}
    \label{tab:res}
    \vspace{-10pt}
\end{table*}
\newline
\textbf{Model Settings:}
For fair comparison with existing benchmarks, we adopted the model configurations and training protocols from PPAL \cite{ppal} for COCO and PASCAL VOC. Our implementation was built on PPAL's \cite{ppal} implementation of MMDetection toolkit \cite{mmdetection}. For RetinaNet experiments the training schedule spanned 26 epochs across all three datasets, with a learning rate decay factor of 0.1 applied at the $20^{th}$ epoch, following \cite{ppal} code. Faster R-CNN hyperparameters followed DivProto \cite{divproto}, while SSD used the settings from MIAL \cite{mial}. ResNet-50 \cite{resnet}  served as the default backbone architecture for RetinaNet and Faster RCNN while VGG-16 \cite{vgg} was used as the base detector for SSD, consistent with \cite{mial}.

While most prior active learning works evaluate large detectors, practical deployments e.g., automotive, commonly rely on edge-friendly models training on large-scale datasets. To assess PAL on such edge models, we experimented with YOLOX-Tiny \cite{yolox} using the official YOLOX toolkit to conduct our experiments. YOLOX-Tiny uses CSPDarkNet \cite{cspdark} for backbone. Default training settings of YOLOX-Tiny were followed except the training epochs were set to 50 and MixUp augmentation was enabled.  Test configuration changes included setting NMS threshold to 0.5 and confidence threshold to 0.3. The training batch size used was 64. We further evaluate PAL's detector-agnostic abilities using  YOLO11s \cite{yolo11}, another edge model, trained and tested on default hyper-parameters except setting training epochs to 50.

The experiments were conducted on a cluster of 4 Tesla V100 DGXS 32GB NVIDIA GPUs, maintaining the hyperparameter settings established in previous works. Model inference hyperparameters used to generate full dataset detections for LIUS were as follows: for RetinaNet and YOLO11s pre-NMS boxes were set to 1000, NMS IoU threshold was set to 0.5 and score threshold was set to 0.3; SSD and Faster R-CNN used their default NMS and score thresholds; YOLOX-Tiny  followed the test-time settings mentioned above. Active learning-specific hyperparameters were set empirically without extensive tuning: $\alpha$ and $\beta$ in \cref{eq:tsi} were set to 0.9 and 0.04 respectively, while $\gamma$ was fixed at 0.02 such that d = 0.1 from \cref{eq:ts} .

\subsection{Results and Analysis}
We compare PAL against state-of-the-art active learning baselines, including PPAL \cite{ppal}, CoreSet \cite{coreset}, CDAL \cite{cdal}, MIAL \cite{mial}, DivProto \cite{divproto}, and LearnLoss \cite{ll}, as well as entropy and random sampling baselines following  \cite{ppal}.
\\
\textbf{RetinaNet Results:}
As shown in \cref{fig:res} (a) and (d), and \cref{tab:res}, we evaluate PAL across three settings: COCO with RetinaNet, PASCAL VOC with RetinaNet, and BDD with RetinaNet. For COCO and PASCAL VOC, we adhere to the training and dataset protocols in PPAL \cite{ppal}. For BDD, the entropy-based sampling baseline computes image uncertainty as the sum of classification entropies over all detected objects, following PPAL \cite{ppal}. Across all three settings, PAL consistently outperforms competing methods: on COCO, PAL yields a +1.4 AP@0.5-0.95 improvement in the last active learning round; on PASCAL VOC, +0.9 mAP@0.5; and on BDD, +1.2 mAP@0.5. On PASCAL VOC, PAL trails PPAL in the first two rounds but overtakes in later rounds as more labelled data is acquired. For VOC, PAL exhibits higher performance variance early on and stabilizes subsequently.

By decoupling active sample selection from model and training internals and relying solely on multiple image-level and inference-time signals, PAL is comparatively better at generalizing effectively across datasets and selects more informative samples. This is particularly evident on COCO and PASCAL VOC, where PAL matches or surpasses PPAL even though PPAL uses, on average, about 20.7\% more annotations across both datasets (see \cref{tab:dd}).  
\begin{table}[!ht]  
  \centering
  \resizebox{0.35\textwidth}{!}{
    \large
  \begin{tabular}{@{}lc@{}}
    \toprule
    Dataset & \% More Annotation (PPAL) \\
    \midrule
    COCO & 18.6\% \\
    PASCAL VOC & 22.8\% \\
    \bottomrule
  \end{tabular}
  }
  \caption{Percentage of additional annotations required by PPAL relative to PAL on COCO and PASCAL VOC using RetinaNet, averaged over all active learning rounds excluding round 1.}
  \label{tab:dd}
  \vspace{-5pt}
\end{table}
\newline
\textbf{SSD and Faster R-CNN experiments:}
In \cref{fig:res} (b), we evaluate PAL on COCO using Faster R-CNN. PAL achieves better results over the previous best PPAL. In \cref{fig:res} (c), we compare PAL with prior approaches on PASCAL VOC using SSD. PAL achieves superior performance to all competing methods, with results in each round at least on par with or better than prior work, demonstrating robustness across architectures, training regimes, and datasets.
\\
\textbf{Results on Other Models:}
  For YOLOX-Tiny, we evaluated PAL on BDD dataset, against random and entropy as baselines. From \cref{tab:res} it can be seen, PAL consistently attains higher mAP, maintaining $\geq+1$ mAP@0.5 improvement over entropy from round 3.
  
  For YOLO11s, PAL is evaluated on COCO  against random  baseline. Like RetinaNet, YOLO11s is trained on COCO with 2\% increments for 5 rounds. Results in \cref{tab:ab1} show that PAL maintains $\geq+1$ AP@0.5-0.95 improvement across AL rounds, illustrating PAL's ability to easily integrate across models while improving performance.
  \begin{table}[!hb]
    \centering
    \resizebox{0.48\textwidth}{!}{
    \large
    \begin{tabular}{|c|c|c|c|c|c|}
        \hline
         \textbf{Model} & \textbf{Method} & \textbf{Round 2} & \textbf{Round 3} & \textbf{Round 4} & \textbf{Round 5} \\
                
       \hline
        \multirow{2}{*}{YOLO11s} & Random & $4.1\pm0.2$ & $6.7\pm0.1$ & $8.7\pm0.1$ & $10.7\pm0.1$  \\
        \cline{2-6}
        & Ours & $5.1\pm0.2$ & $7.9\pm0.2$ & $10.2\pm0.3$ & $12.2\pm0.1$  \\
               
        %\multirow{2}{*}{Data Set 1} & \multirow{2}{*}{A1} & 10.5 & 2.1 & 75 & 1 \\
        %\cline{3-6}
        % & 9.8 & 3.0 & 82 & 0 \\
        %\hline
        %Data Set 2 & A2 & 15.0 & 1.5 & 60 & 1 \\
        \hline
    \end{tabular}
    }
    \caption{AP@0.5-0.95(\%) results for AL rounds on COCO. The 1st round results on 2\% randomly sampled data is $1.7\pm0.1$.}
    
    \label{tab:ab1}  
    \vspace{-5pt}
\end{table}
\\
\textbf{Computational Complexity:} 
After running inference over the entire dataset, PAL’s runtime scales linearly with the number of instances (or the number of images, whichever is larger). PAL's wall-clock time can be further reduced as CLCs in LIUS
can train and run in parallel and GUIDE score calculation can  be parallelized over images. 

\subsection{Ablation Studies}
\textbf{Ablation Settings:} All ablations were conducted on COCO. We started with a 2\% randomly sampled labelled seed set and performed four active learning rounds, each adding an additional 2\% of labelled data selected by the respective strategy. We used RetinaNet to perform ablations with hyperparameters settings following \cref{sec:es}.
\begin{figure*}[!ht]
  \centering
  %\fbox{\rule{0pt}{2in} \rule{0.9\linewidth}{0pt}}
   \includegraphics[width=1.0\linewidth]{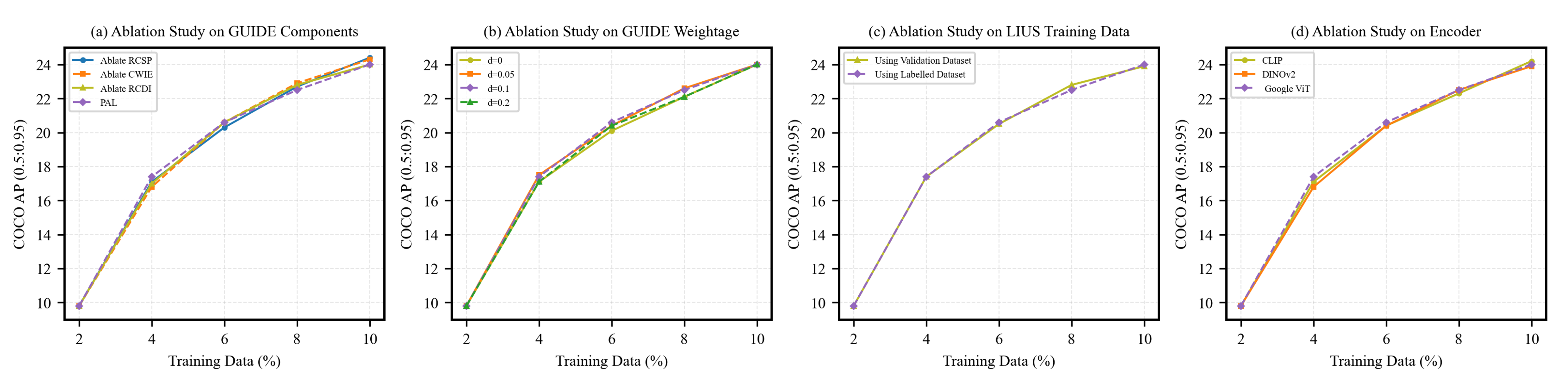}

   \caption{Ablation study on impact of: (a) different components of GUIDE; (b)  GUIDE score contribution to the selection score; (c) labelled dataset used to train logistic classifier models for LIUS; (d) encoder influence; on final performance.}
   \label{fig:ablation}
   \vspace{-10pt}
\end{figure*}
\\
\textbf{GUIDE Components:} We ablated each GUIDE component in turn while keeping the weight of GUIDE score d from \cref{eq:ts} in selection score fixed at 0.1. When removing RCSP, its weight was split equally between CWIE and RCDI; when removing RCDI, its weight was reassigned to CWIE; and when removing CWIE, its weight was reassigned to RCDI. As shown in \cref{fig:ablation} (a), dropping any single component reduces early-round mAP, with the largest drop seen on removing CWIE. Later round results, especially on removing RCSP or CWIE see improvement relative to PAL. This is likely because in later rounds as samples move away from the logistic-classifier decision boundary, the instance score values drop and its variance increases. In such condition the  presence of varying scores of RCSP or CWIE impacts the selection score more than earlier rounds resulting in poorer performance of PAL.  Overall, removing any component of GUIDE lowers early-round performance, indicating that each part of the GUIDE contributes to a better mAP in the low-data iterations.
\begin{figure}[b]
\vspace{-10pt}
  \centering  
    \includegraphics[width=0.95\linewidth]{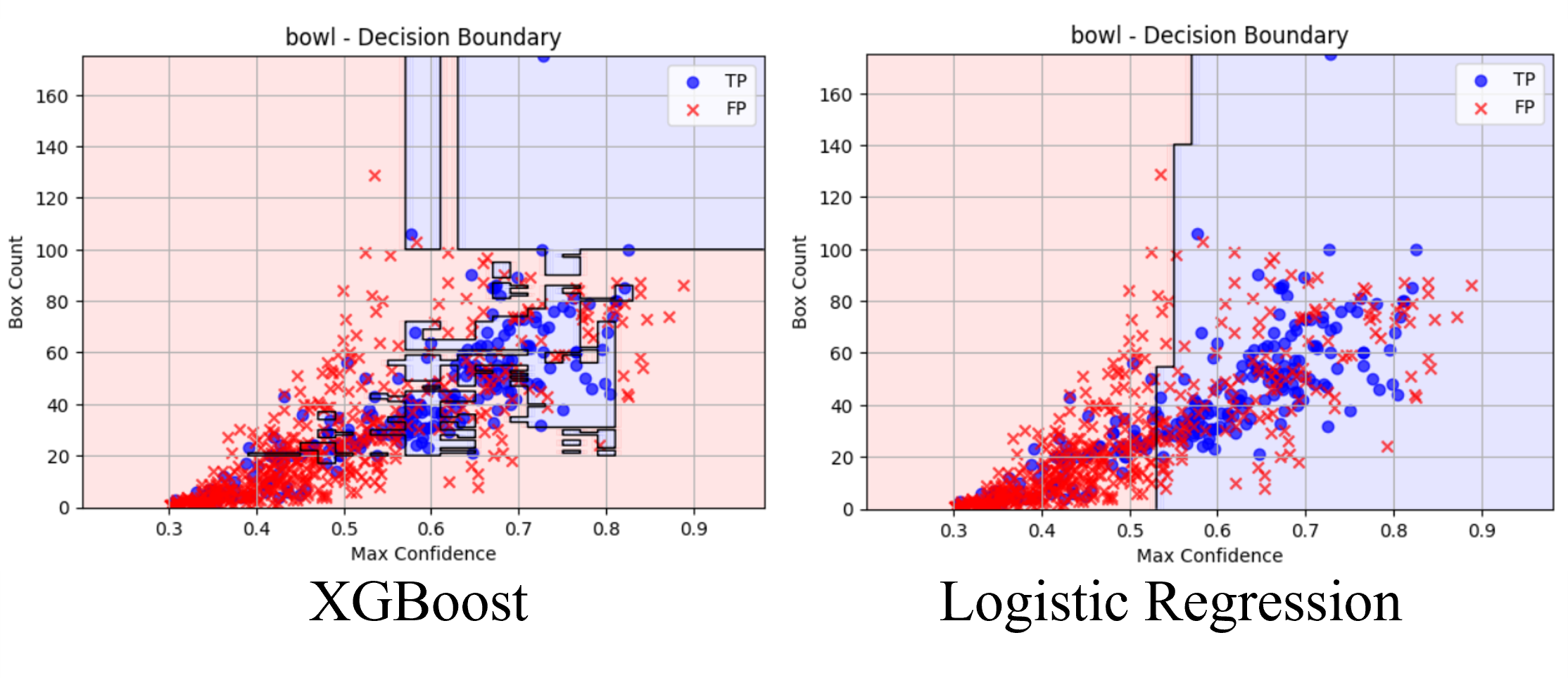}
   \caption{Classification boundary for XGBoost and simple logistic classifier on bowl class from COCO dataset.}
   \label{fig:boundary}
\end{figure}
\\
\textbf{LIUS Only:} We removed the GUIDE scores from the AL selection pipeline and selected data solely by LIUS score. Results in \cref{fig:ablation} (b) where the GUIDE score weight d from \cref{eq:ts} is set to 0 shows a clear degradation in early-round performances without the global component. Although the final-round mAP approaches that of PAL, the early-round gap underscores the importance of diversity, especially when data is small. On COCO, where multiple categories are rare—sometimes too rare to generate per-class logistic classifier model, GUIDE is critical for prioritizing images that benefit the detector most.
\\
\textbf{GUIDE Weightage:} In these experiments, we varied the weightage of LIUS and GUIDE scores in the selection score \cref{eq:ts}. As shown in \cref{fig:ablation} (b), neither increasing nor decreasing the GUIDE weight improves performance; a 0.1 GUIDE weight remains near-optimal under our setup.
\\
\textbf{LIUS Training Data:} PAL trains per-class logistic classifiers on detections from labelled training images, to infer true/false positives (TP/FP). Since the detector has been trained on these images, TP/FP estimates for unlabelled data might be biased. We therefore train the logistic classifiers on a  labelled set separate from training data used for detector training. For these experiments we used validation-set predictions as labelled data for training logistic classifier in every AL round. As shown in \cref{fig:ablation} (c), this yields negligible gains compared to PAL, suggesting that training data is sufficient to compute LIUS.
\\
\textbf{Encoder Influence:} We ablate encoder choice—CLIP \cite{clip}, DINOv2 \cite{dinov2}, and Google ViT \cite{gvit} used in PAL, in \cref{fig:ablation} (d). Google ViT, trained mainly on image classification corpus, yields stronger early-round performance than CLIP known for joint text and image representation and self-supervised DINOv2, known for it's all-purpose features. As the detector sees more data, in later rounds the diversity reduces and encoder choice has less impact on performance.
\\
\textbf{Classifier Choice:} Our results from using more complex classifiers like XGBoost \cite{xgb} over simple logistic classifier, for LIUS calculation, did not show improvement in performance. This was attributed to complex classifiers' tendency to overfit (as seen in \cref{fig:boundary}), especially for low frequency classes. In contrast, the simple logistic classifier captured the TP/FP decision boundary reliably, while avoiding overfitting, and required negligible training time.

%% file: sec/conclusion.tex
\section{Conclusion}
In this work, we introduced Portable Active Learning (PAL), a model-agnostic active learning framework for object detection. PAL performs image selection by synthesizing signals from a logistic-based instance uncertainty scorer (LIUS) and a global uncertainty and image diversity estimator (GUIDE). This instance- and image-aware acquisition strategy consistently improves data quality and yields superior performance over state-of-the-art active learning baselines for detection. By design, PAL is portable and independent of specific model and training pipelines, enabling broad compatibility and strong generalization across detectors, datasets and training regimes.
%that couples an instance-uncertainty scorer (LIUS) with an image-level scorer (GUIDE). GUIDE integrates complementary image-level signals—Class-Weighted Image Entropy (CWIE), Rare-Class driven Diversity Index (RCDI), and Rank-Conditioned Similarity Penalty (RCSP)—to jointly capture uncertainty, diversity, and redundancy. This instance- and image-aware acquisition strategy consistently improves data quality and yields superior performance over state-of-the-art active learning baselines for detection. By design, PAL is plug-and-play and independent of specific training pipelines, enabling broad compatibility and strong generalization across detectors, datasets, and training regimes.

%% file: sec/ac.tex
\section*{Author Contributions}
\textbf{Rashi Sharma} conceptualized the study, designed the methodology, executed the primary experiments and evaluations, drafted the manuscript and secured partial funding.
\\
\textbf{Justin Bersamin} managed the computational environment setup and performed a  subset of the experiments under supervision.
\\
\textbf{Karthikk Subramanian} provided primary funding and performed the literature survey.
\\
All authors reviewed the manuscript and provided feedback. 
\\
\\
\textbf{Acknowledgement.} This work was supported by and carried out at the Panasonic R\&D Center Singapore.  
%Rashi Sharma: Conceptualization, Methodology, Validation, Investigation (Lead), Writing – Original Draft, Project Administration, Funding Acquisition.

%Justin Bersamin: Resources (Environment Setup), Investigation (Supporting).

%Karthikk Subramanian: Funding Acquisition (Lead), Data Curation (Literature Survey).